\documentclass[11pt]{article}
\usepackage[preprint]{acl}

\usepackage{times}
\usepackage{latexsym}
\usepackage[T1]{fontenc}
\usepackage[utf8]{inputenc}
\usepackage{microtype}
\usepackage{amsmath,amssymb}
\usepackage{booktabs,tabularx,multirow}
\usepackage{graphicx}
\usepackage{needspace}
\usepackage{url}

\newcommand{\att}{\mathsf{A}}
\newcommand{\fix}{\mathsf{F}}

\title{Text Grouping Changes Completion-Time Comparisons\\
in Long-Form Simultaneous Speech Translation}

\author{
  Hieu Minh Truong \and Dung D. Le \\
  VinUniversity, Vietnam \\
  \texttt{\{hieu.tm2,dung.ld\}@vinuni.edu.vn}
}
\hypersetup{
  pdftitle={Text Grouping Changes Completion-Time Comparisons in Long-Form Simultaneous Speech Translation},
  pdfauthor={Hieu Minh Truong and Dung D. Le}
}

\begin{document}
\maketitle

\begin{abstract}
Does grouping translation updates before speech synthesis change the measured completion gap between two translators? We compare StreamAtt with its fixed-audio-history baseline on ten English--German talks. For each translation, we keep the intended text and its availability times fixed and vary how updates are grouped for synthesis. Four-second grouping increases StreamAtt's mean completion lead from 106.6 to 229.1 seconds with XTTS-v2, but reduces it from 169.5 to 146.5 seconds with Piper. Both produce shorter audio; different changes in playback idle account for the opposite effects. These directions recur on a separately selected twenty-talk panel. In continuing CosyVoice 3 sessions, grouping instead leaves paired audio unchanged and delays completion on nineteen of twenty blocks. A speech-timing comparison therefore needs to specify how text enters synthesis, not just which synthesizer is used. Recognition and automatic word timing describe the outputs, but do not establish translation accuracy or intelligibility.
\end{abstract}

\section{Introduction}
A speech translator can produce text early and still finish speaking late. Simultaneous translation is commonly evaluated by relating text-emission delay to translation quality \citep{ma2019,ma2020,papi2022,papi2023,papi2024}. Spoken output adds synthesis and serial playback: committed text must become audio, and that audio must be played in order. Work on simultaneous speech-to-speech translation therefore also studies initial waiting, synthesized duration, and output completion \citep{zheng2020,liu2022,kano2023}.

Using the same text-to-speech (TTS) model for two translators does not make their synthesis inputs comparable. One translator may emit many short updates, while another emits fewer, longer ones. The system can synthesize each update separately or group several updates first. Grouping changes both text context and request boundaries; the resulting audio can differ in duration even when the intended words do not change.

\paragraph{Related work.}
\citet{liu2022} study the trade-off between initial waiting and synthesized duration. Related evaluation work examines sensitivity to segmentation, computation accounting, and aggregation \citep{papi2025,xu2025,polak2026}. Long-form speech-output evaluation uses recognition and alignment to examine generated speech \citep{xue2026}, while systems differ in how translated text enters synthesis \citep{djanibekov2026}. Our question is not whether grouping can shorten one system's playback. It is whether grouping changes the completion gap between two translators.

We compare StreamAtt's attention-based audio-history selection with its fixed-audio-history baseline, using the same translation checkpoint \citep{papi2024}. We save each strategy's German text updates and availability times, then replay each stream under different grouping rules. Text stays fixed across grouping conditions within a strategy, not across the two strategies. This control separates the grouping intervention from changes to the saved translation stream.

Four-second grouping widens StreamAtt's mean completion lead with XTTS-v2 but narrows it with Piper. A separate twenty-talk panel reproduces both directions. Duration and playback idle explain where the completion changes occur, but do not isolate an architectural cause. The contribution is this controlled comparison and its implication for evaluation: reporting a TTS model without its text-grouping rule leaves part of the timing comparison unspecified.

\begin{table*}[t]
\centering\small
\begin{tabular}{llrrrrrr}
\toprule
TTS & Translator & Wait (s) & First playback & Audio $D$ & Idle $I$ & Offset $L$ & ASR edits (\%)\\
\midrule
XTTS-v2 & StreamAtt & 0 & 17.44 & 1080.36 & 19.21 & 468.01 & 32.89 \\
 & StreamAtt & 2 & 19.82 & 834.02 & 30.05 & 232.52 & 26.72 \\
 & StreamAtt & 4 & 22.42 & 723.36 & 41.24 & 133.04 & 22.45 \\
 & Baseline & 0 & 17.44 & 1159.30 & 46.85 & 574.59 & 30.90 \\
 & Baseline & 2 & 19.81 & 989.54 & 64.74 & 422.72 & 24.87 \\
 & Baseline & 4 & 22.34 & 919.94 & 73.75 & 362.13 & 24.16 \\
\midrule
Piper & StreamAtt & 0 & 16.85 & 562.51 & 90.14 & 21.09 & 16.45\\
 & StreamAtt & 4 & 21.09 & 488.80 & 150.89 & 8.13 & 13.94\\
 & Baseline & 0 & 16.86 & 692.98 & 129.14 & 190.56 & 17.66\\
 & Baseline & 4 & 21.09 & 642.35 & 143.80 & 154.58 & 16.15\\
\bottomrule
\end{tabular}
\caption{\textbf{Exploratory full-talk results.} Means give each of the ten talks equal weight; all 100 outputs are complete. Times are seconds, and mean source duration is 631.56 seconds. The ASR column compares recognized with intended words, retaining punctuation. Normalized error components appear in Appendix~\ref{app:content}. Compare grouping waits within a TTS configuration: hardware, voice, and frontend differ between backends.}
\label{tab:main}
\end{table*}

\Needspace{6\baselineskip}
\section{Study Design}
\label{sec:design}
\paragraph{Saved streams and controls.}
We separate translation from synthesis. For each talk and strategy, the saved stream contains updates $(x_i,a_i)$: newly committed text and its availability time. A grouping rule combines updates into synthesis input; an input-limit splitter may divide that input into several requests. Within a backend, changing the rule preserves the stream, voice, inference settings, and playback rule. Audio duration, synthesis cost, and playback timing are outcomes. Across backends, request text is shared at matched timer settings, but voice, frontend, and hardware differ.

\paragraph{Talks and translators.}
The exploratory panel contains ten English--German MuST-C talks \citep{digangi2019}, totaling 6,315.6 seconds. They were selected by metadata and inspected before the grouping study; all are retained. A second panel contains twenty additional talks selected before new model outputs, with both histories and immediate/four-second synthesis under both backends. We keep the panels separate.

Both translators use the released StreamAtt checkpoint \citep{papi2023,papi2024}. StreamAtt ($\att$) retains source audio using attention; the baseline ($\fix$) removes audio according to discarded text-history words. On the exploratory panel, StreamAtt emits 4,064 updates containing 14,430 whitespace words; the baseline emits 2,924 updates containing 19,823 words. Both have a median update length of three words. The baseline's 37.4\% greater text length is not evidence of better or worse translation. Settings and source lists are in Appendices~\ref{app:implementation}--\ref{app:confirmation}.

\begin{table*}[t]
\centering\small
\begin{tabular}{llrrrrr}
\toprule
Panel & TTS & $w$ (s) & Gap $\delta(0)$ & Gap $\delta(w)$ & Change $\eta(w)$ & 95\% interval for $\eta$\\
\midrule
Exploratory (10) & XTTS-v2 & 2 & $-106.58$ & $-190.21$ & $-83.63$ & $[-124.22,-44.66]$\\
 & XTTS-v2 & 4 & $-106.58$ & $-229.08$ & $-122.50$ & $[-180.03,-69.92]$\\
 & Piper & 4 & $-169.46$ & $-146.46$ & $+23.01$ & $[7.36,40.99]$\\
\midrule
Confirmation (20) & XTTS-v2 & 4 & $-118.63$ & $-224.92$ & $-106.29$ & $[-148.74,-67.53]$\\
 & Piper & 4 & $-118.23$ & $-84.12$ & $+34.11$ & $[17.52,55.17]$\\
\bottomrule
\end{tabular}
\caption{\textbf{Grouping changes the between-translator completion gap.} Times are seconds; gaps and changes are rounded separately. Negative $\delta$ means StreamAtt finishes earlier. Negative $\eta$ widens its mean lead; positive $\eta$ narrows it in these comparisons. Intervals resample paired talks within each panel. The mean translator ordering does not reverse.}
\label{tab:contrasts}\label{tab:backendinteraction}
\end{table*}

\paragraph{Grouping.}
Immediate dispatch ($w=0$) sends each update separately. With grouping wait $w$, the oldest pending update starts a deadline at $a_i+w$. Updates available by that deadline are joined with one space; later arrivals do not reset the timer. Translation end releases any pending text early. A busy synthesizer delays execution, not group membership, so $w$ is not a bound on total latency. We also test punctuation- and word-count rules with a two-second cap (Appendix~\ref{app:robustness}).

\paragraph{Synthesis and playback.}
XTTS-v2 uses Ana Florence \citep{casanova2024}; Piper uses the German voice \texttt{de\_DE-thorsten-high} \citep{kim2021vits,piperengine,pipervoice}. The original comparison has sixty XTTS outputs at waits 0, 2, and 4 seconds and forty Piper outputs at 0 and 4 seconds. Four seconds was chosen before synthesis; two seconds was a shorter check, not a score-selected setting.

Each request runs separately. Readiness includes measured preparation, inference, and transfer; loading, warmup, logging, and file storage are excluded. Playback is serial at normal speed, without trimming in the primary analysis. We replay saved translation times rather than rerun translation alongside synthesis. XTTS runs on a GPU and Piper on a CPU; this is not a matched-hardware speed comparison.

\paragraph{Continuing synthesis.}
A separate CosyVoice 3 experiment compares immediate and four-second text feeds within one continuing session per block \citep{du2025,funaudiollm}. Twenty paired blocks yield forty sessions. State resets between blocks, not feeds. This tests the feed schedule within a block, not full-talk translator ordering (Appendix~\ref{app:native-input}).

\section{Measurements}
\label{sec:measure}
\paragraph{Completion and its components.}
Primary timing uses all 100 exploratory outputs and all 160 confirmation outputs, without recognition or alignment filtering. Additional policy and seed comparisons are reported separately. Time zero is source start. An audio part $j$ becomes ready at $r_j$, has sample-derived duration $d_j$, and finishes playback at
\begin{equation}
 C_j=\max(r_j,C_{j-1})+d_j,\qquad C_0=0.
 \label{eq:playback}
\end{equation}
For $m$ parts, let $D=\sum_jd_j$ and $I=\sum_j\max(0,r_j-C_{j-1})$. With source duration $S$, the final offset is
\begin{equation}
 L=C_m-S=D+I-S.
 \label{eq:offset}
\end{equation}
Thus $L$ measures completion relative to source end. Silence inside generated audio belongs to $D$; time when playback waits for audio, including the initial wait, belongs to $I$ \citep{liu2022}. Shorter audio need not reduce $L$ if later readiness creates additional idle.

For talk $s$, we measure the completion gap and its change under grouping:
\begin{align}
 \delta_s(w)&=L_s(\att,w)-L_s(\fix,w),\\
 \eta_s(w)&=\delta_s(w)-\delta_s(0).
 \label{eq:interaction}
\end{align}
Negative $\delta$ favors StreamAtt. We calculate each backend separately, giving talks equal weight. Descriptive 95\% intervals use 10,000 paired talk-bootstrap draws \citep{koehn2004}, retaining all conditions of each talk together. They describe source variation within the sampled panel, not new synthesis realizations.

\paragraph{Content and word timing.}
Whisper-large-v3-turbo transcribes the original 100 outputs \citep{radford2023,openaiwhisper}; Qwen3-ASR independently transcribes their immediate/four-second conditions \citep{qwen2026asr}. Word edit rates compare recognition with intended synthesis text. We report the punctuation-retaining rates in Table~\ref{tab:main} and normalized error components in Appendix~\ref{app:content}. These are not source-reference translation scores.

For XTTS, a separate forced aligner estimates when recognized intended words finish playing \citep{qwen2026aligner}. The comparison uses the same unambiguous, validly timed occurrences across all three waits. It covers only a selected subset and does not determine eligibility for completion measurements (Appendix~\ref{app:wordtiming}).

\begin{figure}[t]
\centering
\includegraphics[width=\columnwidth]{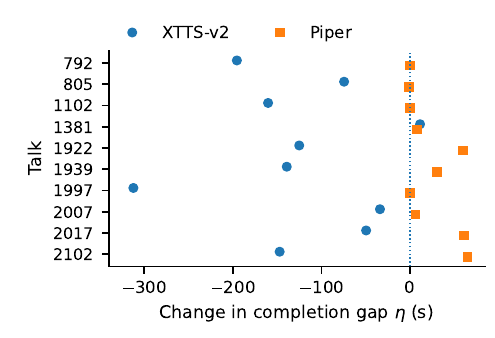}
\caption{Four-second grouping effects on individual exploratory talks. Negative $\eta$ shifts the completion gap toward StreamAtt; positive $\eta$ shifts it toward the baseline. Each backend uses the same ten talks.}
\label{fig:source-effects}
\end{figure}

\section{Results}
\paragraph{The completion gap changes in opposite directions.}
On the exploratory panel, four-second grouping increases StreamAtt's mean lead from 106.58 to 229.08 seconds with XTTS, but reduces it from 169.46 to 146.46 seconds with Piper (Table~\ref{tab:contrasts}). The XTTS effect is negative on nine of ten talks; the Piper effect is positive on seven (Figure~\ref{fig:source-effects}). StreamAtt leads on eight XTTS talks without grouping and all ten with grouping; it leads on all ten Piper talks at both waits. The mean ordering does not reverse.

\paragraph{Separate confirmation.}
On the twenty-talk panel, grouping widens StreamAtt's mean lead from 118.63 to 224.92 seconds with XTTS ($\eta(4)=-106.29$) and narrows it from 118.23 to 84.12 seconds with Piper ($\eta(4)=+34.11$). The exploratory directions recur on 17/20 and 19/20 talks, respectively. Appendix~\ref{app:confirmation} gives the source list, Piper's individual exceptions, and the duration--idle decomposition.

\paragraph{Why shorter audio can narrow a lead.}
Grouping shortens StreamAtt's XTTS audio by 357.00 seconds while adding 22.03 seconds of idle. Most of that saving therefore advances completion. With Piper, StreamAtt saves more audio than the baseline---73.71 versus 50.63 seconds---but adds much more idle: 60.75 versus 14.66 seconds. It finishes 12.96 seconds earlier, while the baseline finishes 35.97 seconds earlier. The baseline gains more at the endpoint, so StreamAtt's lead narrows. This accounting explains where the seconds go, not an architectural cause. On baseline talk 805, changing XTTS grouping from two to four seconds saves 33.408 seconds of audio and adds exactly the same idle: completion is unchanged.

\paragraph{Other grouping rules and synthesis runs.}
The punctuation- and word-count rules retain the four-second effect's direction relative to immediate dispatch. Among non-immediate rules, the mean completion-gap span is 47.34\ seconds for XTTS and 13.57\ seconds for Piper. A punctuation rule differs from a timer with the same two-second cap by $+8.47$ and $-1.91$ seconds, respectively. Paired intervals are $[-0.83,18.97]$ and $[-3.18,-0.72]$ seconds; these direct comparisons were chosen after inspecting the aggregate results. Across three XTTS seeds, $\eta(4)$ ranges from $-127.6$ to $-116.8$ seconds. Removing request-edge silence reduces the effects' magnitudes but preserves their signs (Appendix~\ref{app:robustness}). Across the original run and two additional Piper realizations, the mean interaction ranges from $+23.01$ to $+23.99$ seconds; all three means are positive (Appendix~\ref{app:piper-repeats}).
% Piper's mean interaction ranges from $+22.61$ to $+23.51$ seconds over three realizations.

\paragraph{Starting earlier and finishing earlier differ.}
Four-second grouping delays mean first playback by about five seconds with XTTS and four seconds with Piper. It does not advance every endpoint. Figure~\ref{fig:queue} shows the separate change in ready-audio queue. In continuing CosyVoice sessions, paired waveform samples are identical: grouping instead delays mean completion by 1.49 seconds for StreamAtt blocks and 2.19 seconds for baseline blocks. Nineteen pairs finish later; one finishes 0.038 seconds earlier. Backend, voice, and block context also differ, so this does not isolate persistence alone.

\paragraph{Content behind the timing changes.}
Both recognizers report lower mean total error under grouping, but more XTTS deletion mismatches. Whisper's StreamAtt deletions rise from 612 to 1,313; its insertions fall from 1,954 to 360. These are recognition errors, not verified acoustic omissions. Detected non-speech accounts for only 38.75--43.43\% of StreamAtt's XTTS duration reduction. Without human judgments, we cannot determine how much of the shorter audio reflects fewer repetitions, faster speech, or omitted content. The baseline also produces 37.4\% more intended words; this difference does not establish greater source coverage or translation accuracy.

\section{Conclusion}
Changing how fixed translation updates enter synthesis changes the measured completion gap between translators. The direction differs between XTTS and Piper and recurs on a separate source panel. Evaluations should report grouping and end-of-stream release rules, actual request sizes, audio duration, playback idle, and first playback alongside final completion. These timing differences should not be treated as translation-quality gains: the automatic diagnostics do not establish preserved content.
\label{main-end}

 \section*{Limitations}
The study covers English--German talks, two audio-history strategies from one checkpoint, and one voice per backend. The twenty-talk confirmation does not establish generalization across languages, domains, or voices. XTTS and Piper differ in frontend, implementation, hardware, and voice, preventing architecture-only attribution. Buffered-policy comparisons were selected after inspecting aggregate means; their exploratory intervals do not establish equivalence.

Intended text is fixed within each stream, not across translators, while grouping changes both context and request boundaries. Production replay excludes feedback and device contention. Edge-silence replay holds readiness fixed, and CosyVoice uses restarted short blocks rather than continuous talks.

Without human listening or bilingual adequacy judgments, recognition and alignment remain indirect diagnostics. XTTS timing excludes 7.52\% of alignment units; its validity rule was revised after a timestamp failure without human calibration. The unverified common-word subset covers roughly half of intended words. Source-reference quality and semantic delay remain unreported due to unverified reference coverage and acoustic offsets. For example, talk 805 has annotations ending at 386.66 seconds, but audio continues to 594.55 seconds; the remainder is not assumed silent.

\section*{Ethical Considerations}
Earlier completion does not necessarily imply accurate or complete translation. Our experiments use existing research recordings and publicly released synthesis voices. No new participant recordings or human evaluations were collected. All datasets, models, and voices remain subject to their respective usage conditions.

\paragraph{Use of AI tools.}
The authors used generative AI tools only for code debugging and minor language editing and to improve readability. These tools were not used to generate any scientific content, experimental results, data analyses, or conclusions.

\section*{Acknowledgments}

This research was supported by the VinUniversity
Cross-College Research Grant (Grant ID: VUNI.2324.CC06).

\bibliography{references}

@misc{casanova2024,
      title={XTTS: a Massively Multilingual Zero-Shot Text-to-Speech Model}, 
      author={Edresson Casanova and Kelly Davis and Eren Gölge and Görkem Göknar and Iulian Gulea and Logan Hart and Aya Aljafari and Joshua Meyer and Reuben Morais and Samuel Olayemi and Julian Weber},
      year={2024},
      eprint={2406.04904},
      archivePrefix={arXiv},
      primaryClass={eess.AS},
      url={https://arxiv.org/abs/2406.04904}, 
}

@inproceedings{digangi2019,
    title = "{M}u{ST}-{C}: a {M}ultilingual {S}peech {T}ranslation {C}orpus",
    author = "Di Gangi, Mattia A.  and
      Cattoni, Roldano  and
      Bentivogli, Luisa  and
      Negri, Matteo  and
      Turchi, Marco",
    editor = "Burstein, Jill  and
      Doran, Christy  and
      Solorio, Thamar",
    booktitle = "Proceedings of the 2019 Conference of the North {A}merican Chapter of the Association for Computational Linguistics: Human Language Technologies, Volume 1 (Long and Short Papers)",
    month = jun,
    year = "2019",
    address = "Minneapolis, Minnesota",
    publisher = "Association for Computational Linguistics",
    url = "https://aclanthology.org/N19-1202/",
    doi = "10.18653/v1/N19-1202",
    pages = "2012--2017"
}

@misc{djanibekov2026,
      title={SimulU: Training-free Policy for Long-form Simultaneous Speech-to-Speech Translation}, 
      author={Amirbek Djanibekov and Luisa Bentivogli and Matteo Negri and Sara Papi},
      year={2026},
      eprint={2603.16924},
      archivePrefix={arXiv},
      primaryClass={eess.AS},
      url={https://arxiv.org/abs/2603.16924}, 
}

@misc{du2025,
      title={CosyVoice 3: Towards In-the-wild Speech Generation via Scaling-up and Post-training}, 
      author={Zhihao Du and Changfeng Gao and Yuxuan Wang and Fan Yu and Tianyu Zhao and Hao Wang and Xiang Lv and Hui Wang and Chongjia Ni and Xian Shi and Keyu An and Guanrou Yang and Yabin Li and Yanni Chen and Zhifu Gao and Qian Chen and Yue Gu and Mengzhe Chen and Yafeng Chen and Shiliang Zhang and Wen Wang and Jieping Ye},
      year={2025},
      eprint={2505.17589},
      archivePrefix={arXiv},
      primaryClass={cs.SD},
      url={https://arxiv.org/abs/2505.17589}, 
}

@misc{funaudiollm,
  author = {{FunAudioLLM}},
  title = {{Fun-CosyVoice3-0.5B-2512}},
  year = {n.d.},
  howpublished = {Model card and base checkpoint},
  note = {Revision used: \texttt{29e01c4e}},
  url = {https://huggingface.co/FunAudioLLM/Fun-CosyVoice3-0.5B-2512},
}

@misc{kano2023,
      title={Average Token Delay: A Latency Metric for Simultaneous Translation}, 
      author={Yasumasa Kano and Katsuhito Sudoh and Satoshi Nakamura},
      year={2023},
      eprint={2211.13173},
      archivePrefix={arXiv},
      primaryClass={cs.CL},
      url={https://arxiv.org/abs/2211.13173}, 
}

@inproceedings{koehn2004,
    title = "Statistical Significance Tests for Machine Translation Evaluation",
    author = "Koehn, Philipp",
    editor = "Lin, Dekang  and
      Wu, Dekai",
    booktitle = "Proceedings of the 2004 Conference on Empirical Methods in Natural Language Processing",
    month = jul,
    year = "2004",
    address = "Barcelona, Spain",
    publisher = "Association for Computational Linguistics",
    url = "https://aclanthology.org/W04-3250/",
    pages = "388--395"
}

@misc{liu2022,
      title={From Start to Finish: Latency Reduction Strategies for Incremental Speech Synthesis in Simultaneous Speech-to-Speech Translation}, 
      author={Danni Liu and Changhan Wang and Hongyu Gong and Xutai Ma and Yun Tang and Juan Pino},
      year={2022},
      eprint={2110.08214},
      archivePrefix={arXiv},
      primaryClass={cs.CL},
      url={https://arxiv.org/abs/2110.08214}, 
}

@misc{openaiwhisper,
  author = {{OpenAI}},
  title = {{Whisper large-v3-turbo}},
  year = {n.d.},
  howpublished = {Model card},
  note = {The checkpoint-specific documentation supplements the Whisper paper},
  url = {https://huggingface.co/openai/whisper-large-v3-turbo},
}

@misc{ott2019,
      title={fairseq: A Fast, Extensible Toolkit for Sequence Modeling}, 
      author={Myle Ott and Sergey Edunov and Alexei Baevski and Angela Fan and Sam Gross and Nathan Ng and David Grangier and Michael Auli},
      year={2019},
      eprint={1904.01038},
      archivePrefix={arXiv},
      primaryClass={cs.CL},
      url={https://arxiv.org/abs/1904.01038}, 
}

@inproceedings{papi2023,
  title     = {{AlignAtt: Using Attention-based Audio-Translation Alignments as a Guide for Simultaneous Speech Translation}},
  author    = {Sara Papi and Marco Turchi and Matteo Negri},
  year      = {2023},
  booktitle = {{Interspeech 2023}},
  pages     = {3974--3978},
  doi       = {10.21437/Interspeech.2023-170},
  issn      = {2958-1796},
}

@inproceedings{papi2024,
    title = "{S}tream{A}tt: Direct Streaming Speech-to-Text Translation with Attention-based Audio History Selection",
    author = "Papi, Sara  and
      Gaido, Marco  and
      Negri, Matteo  and
      Bentivogli, Luisa",
    editor = "Ku, Lun-Wei  and
      Martins, Andre  and
      Srikumar, Vivek",
    booktitle = "Proceedings of the 62nd Annual Meeting of the {A}ssociation for {C}omputational {L}inguistics (Volume 1: Long Papers)",
    month = aug,
    year = "2024",
    address = "Bangkok, Thailand",
    publisher = "Association for Computational Linguistics",
    url = "https://aclanthology.org/2024.acl-long.202/",
    doi = "10.18653/v1/2024.acl-long.202",
    pages = "3692--3707"
}

@misc{polak2026,
      title={Better Late Than Never: Meta-Evaluation of Latency Metrics for Simultaneous Speech-to-Text Translation}, 
      author={Peter Polák and Sara Papi and Luisa Bentivogli and Ondřej Bojar},
      year={2026},
      eprint={2509.17349},
      archivePrefix={arXiv},
      primaryClass={cs.CL},
      url={https://arxiv.org/abs/2509.17349}, 
}

@misc{radford2023,
      title={Robust Speech Recognition via Large-Scale Weak Supervision}, 
      author={Alec Radford and Jong Wook Kim and Tao Xu and Greg Brockman and Christine McLeavey and Ilya Sutskever},
      year={2022},
      eprint={2212.04356},
      archivePrefix={arXiv},
      primaryClass={eess.AS},
      url={https://arxiv.org/abs/2212.04356}, 
}

@misc{silero2024,
  author = {{Silero Team}},
  title = {{Silero VAD: Pre-trained enterprise-grade voice activity detector, number detector and language classifier}},
  year = {2024},
  howpublished = {Software repository},
  url = {https://github.com/snakers4/silero-vad},
}

@inproceedings{xu2025,
    title = "{CA}*: Addressing Evaluation Pitfalls in Computation-Aware Latency for Simultaneous Speech Translation",
    author = "Xu, Xi  and
      Xu, Wenda  and
      Ouyang, Siqi  and
      Li, Lei",
    editor = "Chiruzzo, Luis  and
      Ritter, Alan  and
      Wang, Lu",
    booktitle = "Findings of the Association for Computational Linguistics: NAACL 2025",
    month = apr,
    year = "2025",
    address = "Albuquerque, New Mexico",
    publisher = "Association for Computational Linguistics",
    url = "https://aclanthology.org/2025.findings-naacl.393/",
    doi = "10.18653/v1/2025.findings-naacl.393",
    pages = "7077--7082",
    ISBN = "979-8-89176-195-7"
}

@inproceedings{xue2026,
    title = "A Practical Evaluation Method for Long-Form Simultaneous Speech-to-Speech Translation",
    author = "Xue, Yulin  and
      Ouyang, Siqi  and
      Li, Lei",
    editor = "Salesky, Elizabeth  and
      Anastasopoulos, Antonios  and
      Negri, Matteo  and
      Federico, Marcello",
    booktitle = "Proceedings of the 23rd International Conference on Spoken Language Translation ({IWSLT} 2026)",
    month = jul,
    year = "2026",
    address = "San Diego, USA (in-person and online)",
    publisher = "Association for Computational Linguistics",
    url = "https://aclanthology.org/2026.iwslt-1.3/",
    doi = "10.18653/v1/2026.iwslt-1.3",
    pages = "32--39",
    ISBN = "979-8-89176-411-8"
}

@inproceedings{zheng2020,
    title = "Fluent and Low-latency Simultaneous Speech-to-Speech Translation with Self-adaptive Training",
    author = "Zheng, Renjie  and
      Ma, Mingbo  and
      Zheng, Baigong  and
      Liu, Kaibo  and
      Yuan, Jiahong  and
      Church, Kenneth  and
      Huang, Liang",
    editor = "Cohn, Trevor  and
      He, Yulan  and
      Liu, Yang",
    booktitle = "Findings of the Association for Computational Linguistics: EMNLP 2020",
    month = nov,
    year = "2020",
    address = "Online",
    publisher = "Association for Computational Linguistics",
    url = "https://aclanthology.org/2020.findings-emnlp.349/",
    doi = "10.18653/v1/2020.findings-emnlp.349",
    pages = "3928--3937"
}

@misc{qwen2026aligner,
  author = {{Qwen Team}},
  title = {{Qwen3-ForcedAligner-0.6B}: Model Card and Inference Software},
  year = {2026},
  url = {https://huggingface.co/Qwen/Qwen3-ForcedAligner-0.6B},
  note = {Standalone forced-alignment model. Documentation accessed 6 October 2026}
}

@misc{qwen2026asr,
      title={Qwen3-ASR Technical Report}, 
      author={Xian Shi and Xiong Wang and Zhifang Guo and Yongqi Wang and Pei Zhang and Xinyu Zhang and Zishan Guo and Hongkun Hao and Yu Xi and Baosong Yang and Jin Xu and Jingren Zhou and Junyang Lin},
      year={2026},
      eprint={2601.21337},
      archivePrefix={arXiv},
      primaryClass={cs.CL},
      url={https://arxiv.org/abs/2601.21337}, 
}

@misc{piperengine,
  author = {{OHF-Voice}},
  title = {{Piper}: Local Neural Text-to-Speech Engine},
  year = {n.d.},
  url = {https://github.com/OHF-Voice/piper1-gpl},
  note = {Experiment uses package version 1.4.2; full identity retained in the evidence record}
}

@misc{kim2021vits,
      title={Conditional Variational Autoencoder with Adversarial Learning for End-to-End Text-to-Speech}, 
      author={Jaehyeon Kim and Jungil Kong and Juhee Son},
      year={2021},
      eprint={2106.06103},
      archivePrefix={arXiv},
      primaryClass={cs.SD},
      url={https://arxiv.org/abs/2106.06103}, 
}

@misc{pipervoice,
  author = {{Piper Voice Contributors}},
  title = {{de\_DE-thorsten-high}: German Voice Model Card},
  year = {n.d.},
  url = {https://huggingface.co/rhasspy/piper-voices/blob/main/de/de_DE/thorsten/high/MODEL_CARD},
  note = {Pinned voice assets and model-card bytes retained in the experiment}
}

@inproceedings{ma2019,
    title = "{STACL}: Simultaneous Translation with Implicit Anticipation and Controllable Latency using Prefix-to-Prefix Framework",
    author = "Ma, Mingbo  and
      Huang, Liang  and
      Xiong, Hao  and
      Zheng, Renjie  and
      Liu, Kaibo  and
      Zheng, Baigong  and
      Zhang, Chuanqiang  and
      He, Zhongjun  and
      Liu, Hairong  and
      Li, Xing  and
      Wu, Hua  and
      Wang, Haifeng",
    editor = "Korhonen, Anna  and
      Traum, David  and
      M{\`a}rquez, Llu{\'i}s",
    booktitle = "Proceedings of the 57th Annual Meeting of the {A}ssociation for {C}omputational {L}inguistics",
    month = jul,
    year = "2019",
    address = "Florence, Italy",
    publisher = "Association for Computational Linguistics",
    url = "https://aclanthology.org/P19-1289/",
    doi = "10.18653/v1/P19-1289",
    pages = "3025--3036"
}

@inproceedings{ma2020,
    title = "{SIMULEVAL}: An Evaluation Toolkit for Simultaneous Translation",
    author = "Ma, Xutai  and
      Dousti, Mohammad Javad  and
      Wang, Changhan  and
      Gu, Jiatao  and
      Pino, Juan",
    editor = "Liu, Qun  and
      Schlangen, David",
    booktitle = "Proceedings of the 2020 Conference on Empirical Methods in Natural Language Processing: System Demonstrations",
    month = oct,
    year = "2020",
    address = "Online",
    publisher = "Association for Computational Linguistics",
    url = "https://aclanthology.org/2020.emnlp-demos.19/",
    doi = "10.18653/v1/2020.emnlp-demos.19",
    pages = "144--150"
}

@inproceedings{papi2022,
    title = "Over-Generation Cannot Be Rewarded: Length-Adaptive Average Lagging for Simultaneous Speech Translation",
    author = "Papi, Sara  and
      Gaido, Marco  and
      Negri, Matteo  and
      Turchi, Marco",
    editor = "Ive, Julia  and
      Zhang, Ruiqing",
    booktitle = "Proceedings of the Third Workshop on Automatic Simultaneous Translation",
    month = jul,
    year = "2022",
    address = "Online",
    publisher = "Association for Computational Linguistics",
    url = "https://aclanthology.org/2022.autosimtrans-1.2/",
    doi = "10.18653/v1/2022.autosimtrans-1.2",
    pages = "12--17"
}

@article{papi2025,
    title = "How ``Real'' is Your Real-Time Simultaneous Speech-to-Text Translation System?",
    author = "Papi, Sara  and
      Pol{\'a}k, Peter  and
      Mach{\'a}{\v{c}}ek, Dominik  and
      Bojar, Ond{\v{r}}ej",
    journal = "Transactions of the Association for Computational Linguistics",
    volume = "13",
    year = "2025",
    address = "Cambridge, MA",
    publisher = "MIT Press",
    url = "https://aclanthology.org/2025.tacl-1.14/",
    doi = "10.1162/tacl_a_00740",
    pages = "281--313"
}
\appendix
\section{Data and Reproduction Details}
\label{app:implementation}
The exploratory panel contains two development, two tst-COMMON, and six tst-HE talks (Table~\ref{tab:sources}). The same ten sources provide the original 100 full-talk outputs and twenty CosyVoice blocks. History variants, grouping settings, and blocks are not additional independent sources. Selection used metadata and prior-exposure exclusions; the grouping study followed inspection of the resulting translations.

\paragraph{Translation.}
Both strategies use one-second source packets, a commitment setting that withholds a suffix aligned to the most recent $f=6$ encoder frames, attention-layer index 3, and twenty words of textual history. The baseline removes 280~ms of source audio for each discarded history word; it does not reset translation at fixed time intervals. The FBK-fairseq implementation \citep{ott2019,papi2023,papi2024} uses main checkout \texttt{4dde0ea4} and the authors' fixed-audio-history class from \texttt{248135e0}. The twenty exploratory streams contain 6,988 updates and 34,253 whitespace words. Their original availability times are replayed, so upstream computation is not charged again.

\begin{table}[t]
\centering\small
\begin{tabular}{rlr}
\toprule
Talk & Split & Audio (s)\\
\midrule
805 & Dev & 594.547\\
1102 & Common & 421.396\\
792 & Dev & 545.901\\
2017 & HE & 556.536\\
2007 & HE & 551.335\\
1939 & HE & 626.521\\
2102 & HE & 726.251\\
1381 & Common & 731.057\\
1997 & HE & 853.264\\
1922 & HE & 708.789\\
\bottomrule
\end{tabular}
\caption{The ten exploratory MuST-C source recordings. Common and HE denote tst-COMMON and tst-HE. Audio duration is not a manually verified speech-duration measurement.}
\label{tab:sources}
\end{table}

\paragraph{XTTS.}
XTTS-v2 uses the stock Ana Florence voice. The original run sets seed 5242 before each request, with temperature 0.75, length penalty 1.0, repetition penalty 10, top-$k$ 50, top-$p$ 0.85, one beam, sampling enabled, and speed 1.0. The internal sentence splitter is disabled. An external splitter preserves text while respecting 253 characters and the 402-token limit. Inference runs on a partitioned NVIDIA H100 (3g.40gb, 40~GB).

Thirty original full-talk outputs were reused and thirty newly synthesized. Reusing compatible requests preserves their samples and measured synthesis time; that time is included in every replay that uses the request. The reused and new outputs were collected in separate runs. Six predetermined requests reproduced their waveforms on three invocations with the same seed; service-time standard deviations were 0.010--0.056 seconds. This request-level repeatability check is distinct from the full-talk, different-seed comparison in Appendix~\ref{app:robustness}.

\paragraph{Piper.}
Piper 1.4.2 uses the single-speaker 22,050-Hz \texttt{de\_DE-thorsten-high} voice \citep{piperengine,pipervoice}. ONNX runs on the CPU with four intra-operation threads, one inter-operation thread, and sequential execution. Length scale is 1.0, noise scale 0.667, noise-width scale 0.8, normalization is enabled, and volume is 1.0. The original immediate/four-second panel contains 9,377 requests and 10,759 internal sentence outputs. Seventeen punctuation-only requests generate no audio but still incur service cost.

Every Piper request is synthesized afresh. Attempted seed resets did not reproduce sample counts or waveform samples, so audio reuse is disabled. The original full-talk panels contain one realization per condition; Appendix~\ref{app:piper-repeats} reports the separate repetition comparison. 
Four exploratory conditions for talk 792 were collected earlier than the remaining thirty-six, under unchanged settings; the fixed counterbalanced source/condition order is retained.

\paragraph{Timing and uncertainty.}
Request readiness includes frontend processing, inference, and output transfer. Loading, warmup, logging, and file storage are excluded. Playback starts a part only after both its readiness and the preceding part's completion. Full-talk endpoint intervals use 10,000 paired talk-bootstrap draws, seed 1032026, and the recorded order-statistic convention. All conditions of a talk remain together; the exploratory and confirmation panels are not pooled. The intervals are descriptive and are not adjusted for multiple comparisons. Word-timing intervals use their separately recorded convention (Appendix~\ref{app:wordtiming}).

\paragraph{Records.}
The numerical supplement contains translation-release schedules, request and audio-part clocks, completion flags, scoring inputs, software/model revisions, and source-level results. It reconstructs comparisons from saved records; it does not distribute private waveforms or model weights. Incomplete exploratory configurations and stopping records are kept separately.

\section{Source-Level Results and Confirmation}
\label{app:confirmation}\label{app:full}\label{app:allresults}
Tables~\ref{tab:allgaps} and~\ref{tab:pipersources} give every exploratory source's completion contrasts. The XTTS table retains the two-second condition and the individual ordering reversals. Piper's table includes the duration and idle components rather than reporting only a mean interaction.

\begin{table*}[!t]
\centering\small
\begin{tabular}{rrrrrr}
\toprule
Talk & $\delta(0)$ & $\delta(2)$ & $\delta(4)$ & $\eta(2)$ & $\eta(4)$ \\
\midrule
805 & -100.752 & -175.063 & -175.050 & -74.310 & -74.298 \\
1102 & -73.747 & -188.574 & -233.912 & -114.827 & -160.165 \\
792 & +76.843 & -66.650 & -118.590 & -143.493 & -195.433 \\
2017 & -189.004 & -211.120 & -238.417 & -22.116 & -49.413 \\
2007 & -87.943 & -90.048 & -121.696 & -2.105 & -33.753 \\
1939 & -147.164 & -247.972 & -286.235 & -100.808 & -139.071 \\
2102 & -179.145 & -293.291 & -326.248 & -114.146 & -147.103 \\
1381 & -174.269 & -165.853 & -162.750 & +8.416 & +11.519 \\
1997 & +64.410 & -142.197 & -247.929 & -206.607 & -312.339 \\
1922 & -255.040 & -321.316 & -380.021 & -66.276 & -124.981 \\
\bottomrule
\end{tabular}
\caption{All completion gaps and their changes under batching, in seconds. Differences are StreamAtt minus baseline. Negative gaps favor StreamAtt. Gaps and changes are rounded separately.}
\label{tab:allgaps}
\end{table*}

\begin{table*}[!t]
\centering\small
\begin{tabular}{rrrrrrr}
\toprule
Talk & $\delta(0)$ & $\delta(4)$ & $\eta(4)$ & Duration part & Idle part & XTTS $\eta(4)$\\
\midrule
792 & $-30.77$ & $-30.69$ & $+0.08$ & $-30.02$ & $+30.11$ & $-195.43$ \\
805 & $-168.54$ & $-169.29$ & $-0.75$ & $-15.59$ & $+14.84$ & $-74.30$ \\
1102 & $-236.94$ & $-237.01$ & $-0.07$ & $-34.96$ & $+34.88$ & $-160.17$ \\
1381 & $-115.51$ & $-107.62$ & $+7.89$ & $-6.16$ & $+14.06$ & $+11.52$ \\
1922 & $-232.68$ & $-172.73$ & $+59.95$ & $-12.54$ & $+72.49$ & $-124.98$ \\
1939 & $-197.30$ & $-166.47$ & $+30.83$ & $-32.37$ & $+63.20$ & $-139.07$ \\
1997 & $-268.48$ & $-268.62$ & $-0.14$ & $-56.25$ & $+56.11$ & $-312.34$ \\
2007 & $-71.65$ & $-65.46$ & $+6.19$ & $-20.68$ & $+26.86$ & $-33.75$ \\
2017 & $-153.70$ & $-92.73$ & $+60.97$ & $-3.13$ & $+64.11$ & $-49.41$ \\
2102 & $-219.07$ & $-153.94$ & $+65.12$ & $-19.09$ & $+84.21$ & $-147.10$ \\
\bottomrule
\end{tabular}
\caption{All source-level Piper contrasts, in seconds. $\delta$ is StreamAtt minus baseline completion; $\eta=\delta(4)-\delta(0)$. The duration and idle parts sum to $\eta$ up to rounding. Positive $\eta$ narrows StreamAtt's lead. The last column preserves the XTTS contrast on the same source rather than pooling the backends.}
\label{tab:pipersources}
\end{table*}

\subsection{Confirmation panel}
Twenty additional English--German MuST-C talks were selected before new model outputs using seed 20261011. The intended split allocation of four development, four tst-COMMON, and twelve tst-HE talks was changed before synthesis to four, ten, and six, because only six eligible tst-HE talks remained after excluding the exploratory sources. All selected talks were retained. The panel totals 4.55\ hours and spans 251.2--1772.1 seconds. Forty complete translation streams were frozen before synthesis. Both histories and immediate/four-second conditions under both backends give 160 full outputs.

With XTTS, the mean gap changes from $-118.63$ to $-224.92$ seconds: $\eta(4)=-106.29$, with interval $[-148.74,-67.53]$. Seventeen of twenty interactions are negative. The duration component is $-101.77$ seconds and the idle component is $-4.52$ seconds, so most of the interaction comes from the difference in audio-duration savings (Table~\ref{tab:confirm-decomp}). With Piper, the gap changes from $-118.23$ to $-84.12$ seconds: $\eta(4)=+34.11$, with interval $[17.52,55.17]$. Nineteen interactions are positive; the median is $+18.58$ seconds. Both backends retain a mean completion advantage for StreamAtt.

Table~\ref{tab:piper-confirm-source} preserves every Piper confirmation source, including the two talks that favor the baseline at both waits. Talk 1131 has a particularly large positive completion gap; it is retained rather than excluded from the mean. The table describes the timing exception, not its translation or acoustic cause.

\begin{table*}[t]
\centering\small
\setlength{\tabcolsep}{4pt}
\begin{tabular}{rlrrrrrr}
\toprule
Talk & Split & Source (s) & $\delta(0)$ & $\delta(4)$ & $\eta(4)$ & Duration part & Idle part\\
\midrule
783 & Dev & 413.5 & $-72.32$ & $-33.38$ & $+38.94$ & $-0.03$ & $+38.98$ \\
785 & Dev & 1055.9 & $-236.45$ & $-173.84$ & $+62.61$ & $-17.95$ & $+80.56$ \\
769 & Dev & 1318.3 & $-308.70$ & $-254.07$ & $+54.64$ & $-87.89$ & $+142.53$ \\
767 & Dev & 1772.1 & $-396.07$ & $-208.27$ & $+187.80$ & $-32.35$ & $+220.15$ \\
1375 & Common & 974.3 & $-312.88$ & $-252.52$ & $+60.36$ & $-28.89$ & $+89.24$ \\
1161 & Common & 324.8 & $-70.06$ & $-56.06$ & $+14.00$ & $-18.55$ & $+32.55$ \\
1160 & Common & 471.2 & $-79.22$ & $-67.61$ & $+11.61$ & $-0.67$ & $+12.28$ \\
1165 & Common & 553.6 & $-108.91$ & $-98.22$ & $+10.69$ & $-12.52$ & $+23.21$ \\
1378 & Common & 1421.0 & $-199.40$ & $-179.89$ & $+19.51$ & $+18.12$ & $+1.39$ \\
1171 & Common & 251.2 & $-18.27$ & $-17.69$ & $+0.58$ & $-5.20$ & $+5.78$ \\
1346 & Common & 1150.2 & $-99.70$ & $-53.98$ & $+45.73$ & $-31.76$ & $+77.49$ \\
1353 & Common & 381.8 & $+49.34$ & $+77.30$ & $+27.96$ & $-3.53$ & $+31.49$ \\
1144 & Common & 335.8 & $-57.95$ & $-45.54$ & $+12.41$ & $-6.93$ & $+19.34$ \\
1131 & Common & 1041.4 & $+1289.27$ & $+1268.34$ & $-20.94$ & $-32.84$ & $+11.91$ \\
2183 & HE & 1089.1 & $-780.99$ & $-780.83$ & $+0.16$ & $-103.10$ & $+103.26$ \\
2045 & HE & 1117.3 & $-443.71$ & $-355.86$ & $+87.84$ & $-17.45$ & $+105.29$ \\
1932 & HE & 562.2 & $-93.46$ & $-65.97$ & $+27.50$ & $-7.04$ & $+34.53$ \\
2024 & HE & 553.8 & $-44.98$ & $-27.33$ & $+17.65$ & $-22.33$ & $+39.98$ \\
1954 & HE & 996.3 & $-311.95$ & $-298.30$ & $+13.65$ & $-70.25$ & $+83.90$ \\
1961 & HE & 599.2 & $-68.11$ & $-58.62$ & $+9.49$ & $-42.30$ & $+51.79$ \\
\midrule
Mean & & & $-118.23$ & $-84.12$ & $+34.11$ & $-26.17$ & $+60.28$\\
\bottomrule
\end{tabular}
\caption{The twenty-talk confirmation panel and all Piper source-level contrasts. Common and HE denote tst-COMMON and tst-HE. All quantities except talk ID and split are seconds. Negative $\delta$ favors StreamAtt; positive $\eta$ shifts the gap toward the baseline. Components sum to $\eta$ up to rounding.}
\label{tab:piper-confirm-source}\label{tab:confirmation-panel}
\end{table*}

\begin{table}[t]
\centering\small
\begin{tabular}{lrrr}
\toprule
TTS & Duration part & Idle part & $\eta(4)$\\
\midrule
XTTS-v2 & $-101.77$ & $-4.52$ & $-106.29$\\
Piper & $-26.17$ & $+60.28$ & $+34.11$\\
\bottomrule
\end{tabular}
\caption{Confirmation-set mean interaction components, in seconds. Duration and idle account for different parts of the opposite-signed effects.}
\label{tab:confirm-decomp}
\end{table}

\section{Grouping Rules and Timing Sensitivities}
\label{app:robustness}\label{app:additional}\label{app:piper}
\subsection{Practical grouping rules}
All rules act on the same saved streams and preserve the text sequence. The two- and four-second timers start with the oldest pending update. The punctuation rule sends a group once it contains at least three whitespace words and ends in a period, question mark, or exclamation mark (optionally followed by a closing quote or bracket), or after two seconds. The word-count rule sends a group after eight accumulated whitespace words or two seconds. Updates available at the trigger join the group; later arrivals do not reset its timer. Translation end releases any remaining text. Input-limit splitting follows grouping and preserves the joined string.

Every non-immediate rule in Table~\ref{tab:policy-sweep} retains the sign found with the four-second timer relative to immediate dispatch. That does not mean the rules have equal effects. The range of mean gaps among the four non-immediate rules is 47.34\ seconds for XTTS and 13.57\ seconds for Piper.

To compare two buffered policies directly, subtract their mean completion gaps. Punctuation minus the two-second timer gives $+8.47$ seconds for XTTS and $-1.91$ seconds for Piper. Thus differences remain without using immediate dispatch as either member of the contrast. These point contrasts are calculated from the reported rounded means. Paired source-level comparisons are reported in Appendix~\ref{app:policy-paired}.

\begin{table*}[t]
\centering\small
\begin{tabular}{llrrr}
\toprule
TTS & Grouping rule & $\delta(0)$ & $\delta(p)$ & $\eta(p)$\\
\midrule
XTTS-v2 & 2-second timer & $-106.58$ & $-190.21$ & $-83.63$\\
 & Punctuation / 2-second cap & $-106.58$ & $-181.74$ & $-75.16$\\
 & 8 words / 2-second cap & $-106.58$ & $-188.35$ & $-81.77$\\
 & 4-second timer & $-106.58$ & $-229.08$ & $-122.50$\\
\midrule
Piper & 2-second timer & $-169.46$ & $-158.12$ & $+11.34$\\
 & Punctuation / 2-second cap & $-169.46$ & $-160.03$ & $+9.43$\\
 & 8 words / 2-second cap & $-169.46$ & $-154.67$ & $+14.79$\\
 & 4-second timer & $-169.46$ & $-146.46$ & $+23.01$\\
\bottomrule
\end{tabular}
\caption{Exploratory mean completion gaps under four grouping rules, in seconds. Immediate dispatch is the reference for $\eta(p)=\delta(p)-\delta(0)$. Comparisons between buffered rules use differences of $\delta(p)$ instead.}
\label{tab:policy-sweep}
\end{table*}

\subsection{Direct buffered-policy comparisons}
\label{app:policy-paired}
The direct comparison uses the punctuation rule and a timer with the same two-second cap. Group membership, input-limit splitting, and dispatch times are reconstructed from the saved streams. A punctuation trigger accounts for 3.39\% of StreamAtt groups and 2.38\% of baseline groups; end-of-translation release accounts for 0.52\% and 0.63\%. The remaining groups reach the timer.

We resample paired talks, retaining both histories and both policies in each draw. These direct contrasts were selected after inspecting the aggregate policy means, so the analysis is exploratory.

\begin{table}[t]
\centering\small
\setlength{\tabcolsep}{3.5pt}
\begin{tabular}{lrrr}
\toprule
TTS & Mean (s) & 95\% interval & Positive talks\\
\midrule
XTTS & $+8.47$ & $[-0.83,18.97]$ & 6/10\\
Piper & $-1.91$ & $[-3.18,-0.72]$ & 2/10\\
\bottomrule
\end{tabular}
\caption{Punctuation-minus-two-second-timer completion-gap contrasts. Point estimates follow from the policy means in Table~\ref{tab:policy-sweep}. Intervals and direction counts use paired source contrasts.}
\label{tab:policy-paired}
\end{table}

The XTTS paired interval is $[-0.83,18.97]$ seconds, with a positive difference on 6/10 talks; the Piper interval is $[-3.18,-0.72]$, with 2/10 positive talks. Most groups reach the common timer cap, so this contrast compares two similar dispatch rules. It does not establish that all reasonable buffered policies are interchangeable.

\subsection{XTTS synthesis seeds}
% Supplied seed-level means: 5242 = -122.50; 5243 = -116.80; 5244 = -127.60 s.
The full-talk comparison uses seeds 5242, 5243, and 5244 on the ten exploratory talks, both histories, and immediate/four-second grouping. The original seed-5242 outputs are retained; all non-seed settings remain fixed. The three seed-level mean effects are negative, ranging from $-127.6$ to $-116.8$ seconds. The original effect is $-122.50$ seconds. The seed is reset before each synthesis request; seed-level records are retained in the numerical supplement. This checks the sign and scale of the observed XTTS interaction, not every possible synthesis realization.

\subsection{Piper realization variability}
\label{app:piper-repeats}
% The ten exploratory talks, both histories, and immediate/four-second grouping are synthesized under three independent realizations per condition, giving 120\ outputs including the compatible original run. Voice and inference settings are unchanged, execution order is counterbalanced, and completed outcomes are retained. Realization identifiers are used instead of claiming controlled seeds. Technical execution failures are distinguished from normally completed outputs; an unfavorable timing or recognition score is not an execution failure.

% We compute each talk's interaction for every realization, then summarize the panel means and the within-talk standard deviations across realizations. The mean interaction ranges from $+22.61$ to $+23.51$ seconds, with 3/3\ positive panel means. The median within-talk standard deviation is 0.81\ seconds and the maximum is 1.82\ seconds. Between seven and eight talks have a positive interaction in each realization. Source effects near zero can change sign even when the panel means agree. The realizations are not additional independent source talks.
The comparison combines forty archived Piper outputs with eighty newly synthesized outputs: the ten exploratory talks, both histories, and immediate/four-second grouping in each of three realizations. Voice, inference settings, and CPU model are unchanged. The two new runs use counterbalanced condition orders; the original run is retained as a historical realization rather than treated as counterbalanced. Realization identifiers do not imply controlled or independent random seeds. All completed outcomes are retained.

The mean interaction is $+23.01$ seconds in the original run and $+23.71$ and $+23.99$ seconds in the two new runs. The respective numbers of talks with positive interactions are 7/10, 8/10, and 9/10. The two new runs have 95\% paired-talk bootstrap intervals of $[7.97,41.65]$ and $[8.86,40.94]$ seconds. The average narrowing direction therefore recurs in both repetitions.

For each talk, we compute the sample standard deviation of its interaction across the three realizations. The median of these standard deviations is 0.96 seconds and the maximum is 6.37 seconds. Talk 1381 has the largest range, from $+7.89$ to $+20.16$ seconds. Some near-zero source effects change sign across runs. These repetitions describe variability on the same ten talks; they are not additional independent sources or evidence of equivalence between realizations.

\subsection{Request-edge silence}
The edge-silence replay keeps each request's measured readiness and changes only its sample duration. Silero probabilities \citep{silero2024} are evaluated at thresholds 0.3, 0.5, and 0.7. Leading and trailing material outside a 100-ms speech guard is removed; interior pauses and request order are unchanged. Playback is recomputed from the modified durations.

Across thresholds, the interaction ranges from $-96.37$ to $-87.58$ seconds for XTTS and from $+16.94$ to $+19.72$ seconds for Piper (Table~\ref{tab:edge-trim}). At threshold 0.5, the effects are $-91.76$ and $+18.43$ seconds, respectively. Removing the detected request edges reduces the magnitude but preserves the negative XTTS and positive Piper directions. Because readiness is held fixed, this is a retrospective sensitivity analysis, not a timed online trimming implementation or a content-preservation test.

\begin{table}[t]
\centering\small
\begin{tabular}{rrr}
\toprule
Threshold & XTTS $\eta(4)$ & Piper $\eta(4)$\\
\midrule
0.3 & $-96.37$ & $+19.72$\\
0.5 & $-91.76$ & $+18.43$\\
0.7 & $-87.58$ & $+16.94$\\
\bottomrule
\end{tabular}
\caption{Mean grouping effects after request-edge silence removal, in seconds. The primary analysis uses untrimmed audio.}
\label{tab:edge-trim}
\end{table}

\subsection{Piper readiness checks}
In the original panel, Piper grouping shortens all twenty waveforms, but advances completion on only eight StreamAtt and seven baseline talks. The mean interaction is $+23.01$ seconds, median $+7.04$, and range $[-0.75,65.12]$. The duration and idle components are $-23.08$ and $+46.09$ seconds.

A sentence-release replay makes each internal sentence available at its recorded readiness, instead of waiting for the whole request. A zero-service replay removes downstream computation time. Both retain the same waveform samples, request grouping, and original text times. Their interactions are $+23.22$ and $+23.50$ seconds, respectively. Neither completed-request buffering nor measured CPU service alone explains the positive sign. Sentence-level audio release is not a different rule for grouping text before synthesis.

The source-paired difference between Piper's and XTTS's four-second interactions is $+145.51$ seconds, with interval $[90.83,198.50]$. Nine sources are positive; talk 1381 is the exception. This comparison includes voice, frontend, implementation, and device differences.

\subsection{Duration savings and the final endpoint}
For nonnegative readiness times, repeated substitution in Equation~\ref{eq:playback} gives
\begin{equation}
 C_m=\max_{1\leq k\leq m}\left(r_k+\sum_{j=k}^{m}d_j\right).
 \label{eq:critical-part}
\end{equation}
Each candidate combines one readiness event with the audio that follows it. A maximizing $k$ identifies the event that limits completion. Duration saved before that event can become idle rather than advance the endpoint. The baseline talk-805 example in Section~4 shows this cancellation with the reported durations. The identity describes playback accounting; it does not predict audio duration from text or isolate an acoustic mechanism.

\section{Request Structure and Content Diagnostics}
\label{app:content}\label{app:acousticcloseout}
\subsection{Updates and requests}
Whitespace-word counts include standalone punctuation and music symbols. An update is not necessarily one synthesis request: input-limit splitting can create several requests. Table~\ref{tab:updates} describes the text streams; Table~\ref{tab:requestcounts} counts actual XTTS request occurrences after splitting. Reused requests are counted and timed each time their audio enters a replay. A text-assembly check recovers the stored canonical text in all sixty streams across the exploratory and confirmation panels. Packet-to-update agreement is also verified for all forty confirmation streams; the corresponding check is unavailable for the twenty exploratory streams. These checks concern stream assembly, not translation adequacy, and do not explain the baseline's greater intended-text length.

\begin{table}[t]
\centering\small
\begin{tabular}{lrr}
\toprule
Update statistic & StreamAtt & Baseline \\
\midrule
Number of updates & 4064 & 2924 \\
Whitespace words & 14430 & 19823 \\
Median words & 3 & 3 \\
90th-percentile words & 6 & 10 \\
At most two words (\%) & 43.36 & 40.97 \\
Final punctuation (\%) & 1.77 & 1.44 \\
\bottomrule
\end{tabular}
\caption{Text-update lengths and punctuation, pooled over ten talks per system. An update can be split into several synthesis requests. Terminal punctuation does not label a prosodic boundary.}
\label{tab:updates}
\end{table}

\begin{table}[t]
\centering\small
\begin{tabular}{lrrr}\toprule
History & 0 s & 2 s & 4 s \\
\midrule
StreamAtt & 4,068 & 1,894 & 1,149 \\
Baseline & 3,064 & 1,586 & 1,096 \\
\bottomrule\end{tabular}
\caption{Actual XTTS request occurrences across ten recordings. Reused requests count once per use; these are not counts of unique model invocations. Input-limit splitting can create more than one request per update.}
\label{tab:requestcounts}
\end{table}

Four-second grouping removes 2,919 StreamAtt requests and 1,968 baseline requests. Total waveform savings divided by removed requests are 1.223 and 1.216 seconds, respectively; the zero-to-two-second ratios are 1.133 and 1.149. These are descriptive ratios, not a fixed cost per boundary: context, punctuation, request length, and restarts change together.

\subsection{Recognition and scoring}
Whisper receives 16-kHz audio, a German transcription prefix, and no intended text. Four prefix positions leave up to 444 new decoder positions. Capacity-limited fifteen-second windows are split into 7.5-second halves and, when needed, 3.75-second quarters. Transcripts are not selected by reference score or completed from intended text.

The Table~\ref{tab:main} word metric lowercases whitespace tokens but retains punctuation. The normalized breakdown lowercases each token, retains Unicode letters/numbers and ASCII apostrophes, and discards tokens that become empty. Counts are pooled for the error breakdown, whereas mean rates give talks equal weight. Neither statistic measures adequacy against the English source.

Normalized Whisper mean word edit rates decrease from 23.05\% to 14.87\% for StreamAtt--XTTS and from 22.38\% to 17.73\% for baseline--XTTS. Piper's rates decrease from 7.65\% to 6.60\% and from 9.92\% to 9.23\%. However, XTTS deletions increase on nine StreamAtt talks and six baseline talks. Normalized total error worsens on StreamAtt talk 1102 and baseline talk 2007. Table~\ref{tab:editcomponents} retains these adverse error components; a deletion mismatch is not a verified omitted word.

\begin{table*}[!t]
\centering\small
\begin{tabular}{llrrrr}
\toprule
Backend & History & Intended words & Substitutions & Insertions & Deletions\\
\midrule
XTTS-v2 & StreamAtt & 14,336 & $765\to446$ & $1{,}954\to360$ & $612\to1{,}313$\\
 & Baseline & 19,766 & $968\to602$ & $1,452\to458$ & $2,038\to2,441$\\
Piper & StreamAtt & 14,336 & $784\to640$ & $154\to144$ & $163\to168$\\
 & Baseline & 19,766 & $1,108\to1,009$ & $231\to255$ & $506\to432$\\
\bottomrule
\end{tabular}
\caption{Pooled punctuation-normalized edits across all ten recordings, immediate $\to$ four-second grouping. Intended words are fixed within a history across waits and backends. Counts are pooled for the error breakdown; mean percentages elsewhere weight recordings equally. A deletion is an ASR mismatch, not a confirmed acoustic omission.}
\label{tab:editcomponents}
\end{table*}

\paragraph{Full-output Qwen recognition.}
Qwen3-ASR-0.6B \citep{qwen2026asr} independently transcribes the exploratory immediate/four-second outputs without the intended text. Audio preparation, German language specification, tokenization, and scoring normalization are fixed across conditions. Table~\ref{tab:qwen-full} reports mean word edit rates and pooled deletion counts. Like Whisper, Qwen reports lower mean total error in all four backend--history comparisons but higher XTTS deletion counts. Agreement between recognizers reduces dependence on one recognizer; it does not establish what listeners hear.

\begin{table}[t]
\centering\small
\setlength{\tabcolsep}{2.5pt}
\begin{tabular}{llrr}
\toprule
TTS & Translator & ASR edits (\%) & Deletions\\
\midrule
XTTS & StreamAtt & $30.70\to21.60$ & $738\to1{,}186$\\
 & Baseline & $29.40\to23.20$ & $1{,}984\to2{,}268$\\
Piper & StreamAtt & $15.80\to13.20$ & $181\to190$\\
 & Baseline & $17.10\to15.50$ & $487\to451$\\
\bottomrule
\end{tabular}
\caption{Full-output Qwen recognition, immediate $\to$ four-second grouping. Rates weight talks equally; deletion counts are pooled. These are recognition mismatches, not verified acoustic omissions.}
\label{tab:qwen-full}
\end{table}

\paragraph{Selected-clip comparison.}
An earlier input-selected XTTS sample contains forty immediate/grouped pairs, with 15--45 lexical words per block and matching complete request spans. Selection seed 20261007 gives eighty clip IDs covering seventy-three distinct stored waveforms, including seven identical-audio pairs. Whisper and Qwen independently transcribe disjoint 7.5-second windows, with German specified and no intended text; all 160 clip--recognizer results complete. Both report fewer insertions in this sample, while StreamAtt deletions decrease and baseline deletions increase slightly. Table~\ref{tab:clip-edits} gives the pooled error counts. Different selection and recognition context mean this clip result does not replace the full-output deletion increase.

\begin{table}[t]
\centering\small
\begin{tabular}{llrr}
\toprule
History & Recognizer & Insertions & Deletions\\
\midrule
StreamAtt & Whisper & $130\to25$ & $4\to2$\\
StreamAtt & Qwen ASR & $159\to15$ & $2\to0$\\
Baseline & Whisper & $51\to20$ & $7\to9$\\
Baseline & Qwen ASR & $58\to18$ & $1\to2$\\
\bottomrule
\end{tabular}
\caption{Selected-clip edit counts, immediate $\to$ four-second grouping. Each row pools twenty clips per wait from ten recordings. Recognition context, windowing, and selected text coverage differ from the full-recording diagnostic; these counts do not replace its deletion increase.}
\label{tab:clip-edits}
\end{table}

\subsection{Speech activity and word duration}
Silero speech probabilities \citep{silero2024} are calculated on consecutive 512-sample frames at 16~kHz, resetting state between waveforms. Thresholds 0.3, 0.5, and 0.7 use the same probabilities without padding or gap merging and retain the actual final-frame duration. The complete analysis includes immediate/four-second StreamAtt--XTTS outputs for all ten talks; other coverage is partial and is not imputed.

\begin{table}[t]
\centering\small
\begin{tabular}{rrrr}
\toprule
Threshold & Non-speech (s) & Speech (s) & Share (\%) \\
\midrule
0.3 & 138.35 & 218.65 & 38.75 \\
0.5 & 146.18 & 210.82 & 40.95 \\
0.7 & 155.04 & 201.96 & 43.43 \\
\bottomrule
\end{tabular}
\caption{Mean waveform reduction under four-second XTTS grouping for StreamAtt, partitioned by speech activity. Times are seconds; all ten immediate/grouped pairs are included. The final column divides summed non-speech reduction by summed total reduction, not a mean of per-talk percentages. Detector-classified speech does not establish preserved linguistic content.}
\label{tab:vad}
\end{table}

Detected non-speech explains 38.75--43.43\% of the waveform reduction. Detector-classified speech does not establish preserved linguistic content. Repetition, omission, speaking rate, and detector or recognition error may contribute to the remaining reduction.

A separate word-duration analysis uses unambiguous matches to identical intended occurrences, excluding invalid, ambiguous, and request/window-crossing intervals. Duration is compact-audio end minus start, not playback delay. Request-final words that become request-interior shorten by a mean 0.132 seconds for StreamAtt (1,637 occurrences) and 0.131 seconds for the baseline (1,113). Words interior in both conditions shorten by 0.044 seconds (5,219) and 0.017 seconds (8,843). Means give each talk equal weight; all ten talks contribute. Overlapping intervals are counted once. These estimates are consistent with boundary lengthening, but matching and automatic timestamps do not identify a causal boundary effect or explain all speech-classified audio. The aligner's 80-ms bins are resolution, not measured accuracy.

\section{Automatic Word-Timing Assessment}
\label{app:wordtiming}
This supporting analysis uses the original sixty XTTS outputs. It asks when recognizable intended words finish playing after their text becomes available, not when the source first supports their meaning or when a listener understands them. It does not filter any full-talk completion result.

\subsection{Alignment and playback coordinates}
Qwen3-ForcedAligner-0.6B \citep{qwen2026aligner} aligns each retained Whisper window to that same transcript. It is not a second recognizer. Inference uses German, 16-kHz audio, bfloat16, scaled dot-product attention, and batches of four. Recorded versions are \texttt{qwen-asr} 0.0.6, PyTorch 2.9.1, Transformers 4.57.6, and NumPy 2.2.6. All 3,860 windows are processed, with 3,845 distinct prediction inputs. Alignment is offline and does not enter the production clock.

The compact waveform concatenates generated audio but omits playback idle. Let $u$ be a predicted compact-audio coordinate, $q_j$ the compact start of part $j$, and $p_j=\max(r_j,C_{j-1})$ its playback start. A coordinate in part $j$ maps to
\begin{equation}
 t_{\mathrm{play}}(u)=p_j+(u-q_j).
 \label{eq:wordmap}
\end{equation}
The recognition-window offset is added first. At a part join, word ends belong to the preceding part and word starts to the following part. Crossing intervals use their mapped endpoints. This retains later playback gaps rather than adding only the first-playback delay. Mapping uses 12,857 request parts and 90,859 retained timing units.

\subsection{Matching and added delay}
The denominator counts original whitespace words, including punctuation-only tokens. Matching units retain letters, numbers, and apostrophes, omit punctuation, and ignore case while preserving original character offsets. Punctuation-only tokens have no timed unit. Each intended word inherits the latest recorded availability of its characters.

A unit-cost edit alignment compares intended and recognized units. An occurrence is accepted only when every optimal edit path assigns the same exact recognized match and none leaves it unmatched. Timestamps never resolve a repeated-word ambiguity. Every unit in an accepted word needs an unambiguous match and a valid time; its end is the latest matched-unit end. Substitutions, omissions, ambiguous matches, and invalid times remain unknown, not zero delay.

Let $M_{sh}$ contain the intended occurrences accepted at all three waits for talk $s$ and history $h$. If $e_{shi}(w)$ is the predicted playback end and $a_{shi}$ the fixed text availability, added delay is $\ell_{shi}(w)=e_{shi}(w)-a_{shi}$. We compare within-talk quantiles:
\begin{align}
 \Delta^{(q)}_{sh}(w)
 &=Q_q\bigl(\{\ell_{shi}(w):i\in M_{sh}\}\bigr)\notag\\
 &\quad-Q_q\bigl(\{\ell_{shi}(0):i\in M_{sh}\}\bigr),
 \label{eq:wordcontrast}
\end{align}
for $q\in\{0.5,0.9\}$, then average over talks. These are differences of quantiles, not quantiles of paired word differences. Occurrences are matched across waits within a history, not between the translators' different texts.

\begin{table*}[!t]
\centering\small
\begin{tabular}{lrrrrrl}
\toprule
History & Wait (s) & $\Delta$ median & 95\% interval & $\Delta$ p90 & 95\% interval & Lower median \\
\midrule
StreamAtt & 2 & -119.03 & [-145.18, -95.48] & -219.00 & [-261.47, -177.58] & 10/10 \\
StreamAtt & 4 & -173.56 & [-207.73, -140.76] & -314.88 & [-377.34, -252.99] & 10/10 \\
Baseline & 2 & -72.70 & [-106.75, -41.46] & -145.13 & [-204.97, -88.56] & 9/10 \\
Baseline & 4 & -105.36 & [-158.36, -57.25] & -203.06 & [-289.41, -120.88] & 9/10 \\
\bottomrule
\end{tabular}
\caption{Automatically estimated added delay on the common three-setting word subset. Entries are equal-talk means of within-talk quantile changes from separate requests; all times and intervals are seconds. Negative values mean less delay. The final column counts talks with a lower median, not individual words. Every row uses all ten talks. Intervals resample talks and do not include recognition or timestamp error.}
\label{tab:wordtiming}
\end{table*}

\subsection{Coverage and uncertainty}
The common subset contains 8,508 of 14,430 StreamAtt words (58.96\%) and 10,554 of 19,823 baseline words (53.24\%). Equal-talk mean coverage is 58.51\% and 53.91\%, respectively. Per-talk coverage and changes are retained in the numerical supplement. The same subset improves comparability across waits but may not represent excluded words; its coverage gap is not an omission rate.

Table~\ref{tab:wordtiming} uses 10,000 paired talk-bootstrap draws, seed 20261005, and NumPy's default linear quantiles. All three outputs and the common subset remain together. Intervals do not include recognizer or timestamp error. The numerical supplement retains the original analysis by text-availability interval, including words released after source end; no final speech tail was discarded.

\subsection{Invalid timestamps and sensitivity}
After an initial timestamp failure, the rule was changed to retain isolated invalid units as unknown rather than reject the whole panel. This post-observation change was not independently calibrated. Invalid times were not clipped or regenerated to obtain usable values.

Of 98,243 recognized units, 90,859 retain times and 7,384 (7.52\%) remain unknown. Overlapping flags identify 6,313 units changed by the aligner's nonmonotonic-timestamp postprocessor, 4,395 zero-duration units, 716 out-of-window units, and 64 duplicates. These counts are not disjoint. A single-unit window with an unknown time remains in coverage. Unknown units and the larger common-word coverage gap have different denominators; the latter also excludes words without unambiguous recognition at every wait. The accepted transcripts and timestamps have not been independently checked by listeners. Matching an intended word does not establish its source faithfulness or confirm that it was spoken.

A post-hoc conservative check excludes every common word whose recognition window has any flagged timestamp in any output. It retains only 701 StreamAtt words (4.86\%) and 752 baseline words (3.79\%). Mean within-talk median changes are $-133.31/-195.28$ seconds for StreamAtt and $-87.17/-124.27$ for the baseline at two/four seconds. Lower-median counts remain 10/10 and 9/10. The small retained subset and unchanged signs do not validate timestamp accuracy.

\section{Continuing Text-Input Synthesis}
\label{app:native-input}
Twenty blocks were selected before this experiment, one from each exploratory talk--history stream. They contain 64--151 normalized lexical words spanning 14.43--32.66 seconds of original text availability. Each block has an immediate-feed and a four-second-feed condition. Both start with fresh synthesis state and an empty playback queue, then preserve state across feeds. The grouped condition uses the oldest-update deadline and releases pending text at block end. Blocks are not source-aligned between histories; only conditions on the same block are compared.

\paragraph{Configuration and clock.}
Fun-CosyVoice3-0.5B-2512 uses model revision \texttt{29e01c4e} and inference code \texttt{074ca6dc} \citep{du2025,funaudiollm}. A timed Python text generator enters the native bistream interface. The fixed reference voice and prompt are identified in the numerical supplement. Settings are FP32, seed 5242 per session, speed 1.0, initial 25-token acoustic hop, and disabled frontend normalization. Native text/acoustic postprocessing is unchanged; independently synthesized fragments are not substituted for continuing generation.

Controls with two feeds fifteen seconds apart verify output before the final feed, retained state, natural completion, and parity of instrumented and uninstrumented generation. All forty sessions reach native end of sequence and drain fully. The 180-second wall-time and 240-second emitted-audio limits do not censor a session. These checks establish execution behavior, not listening quality.

Readiness is measured from the first original block update and includes feed waits and the instrumented host loop; loading and prompt preparation precede that clock. Completion relative to the last original text update is
\begin{equation}
 B_b=C_{\mathrm{last},b}-a_{\mathrm{last},b}
     =D_b+I_b-a_{\mathrm{last},b}.
 \label{eq:native-offset}
\end{equation}
This is not a full-talk source-end offset.

\begin{table}[t]
\centering\small
\begin{tabular}{@{}lrrrr@{}}
\toprule
History & $\Delta$ first & $\Delta D$ & $\Delta B$ & Later end\\
\midrule
StreamAtt & $+3.261$ & 0 & $+1.486$ & 9/10\\
Baseline & $+3.066$ & 0 & $+2.194$ & 10/10\\
\bottomrule
\end{tabular}
\caption{Native CosyVoice paired-block changes, four-second minus immediate feeding. Means are seconds over ten blocks per row. Each pair has the same waveform samples; all forty sessions complete. $B$ is completion after the block's final original text update. These within-block contrasts are not full-talk translator rankings.}
\label{tab:nativeblocks}
\end{table}

\paragraph{Paired outcomes.}
Each pair has identical flattened text-token sequences, committed speech units, recognized strings, and waveform samples. A check of the archived token records confirms sequence equality in all twenty pairs, although the serialized batch dictionaries differ in every pair. The supplement retains the sample-identity checks and part/feed traces. Normalized mean edit rates are unchanged at 4.84\% for StreamAtt blocks and 13.63\% for baseline blocks.

Grouping changes $B_b$ by a mean $+1.486$ seconds for StreamAtt blocks (nine later; range $[-0.038,3.722]$) and $+2.194$ for baseline blocks (ten later; range $[0.084,3.830]$). Source-bootstrap intervals are $[0.849,2.193]$ and $[1.416,2.961]$, using 10,000 draws, seed 1032026, and the endpoint order-statistic convention. They exclude fresh-seed and execution-time variability. The one 38-ms earlier endpoint is not treated as a reliable speedup.

With unchanged duration, idle accounts for the endpoint differences. The large XTTS shortening is absent in this continuing-session block protocol. Backend, voice prompt, frontend, and context also differ, so persistence alone is not isolated; these blocks do not establish full-talk performance.

% Restored original ready-audio queue figure and its existing measurements.
\begin{figure*}[t]
\centering
\includegraphics[width=0.485\textwidth]{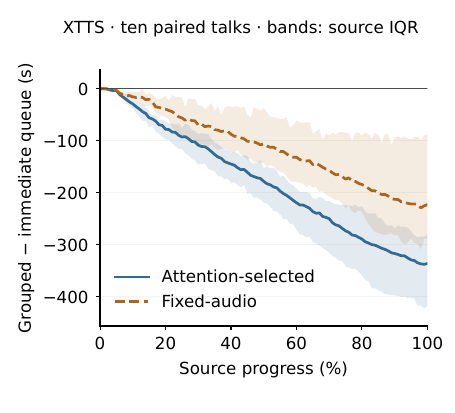}\hfill
\includegraphics[width=0.485\textwidth]{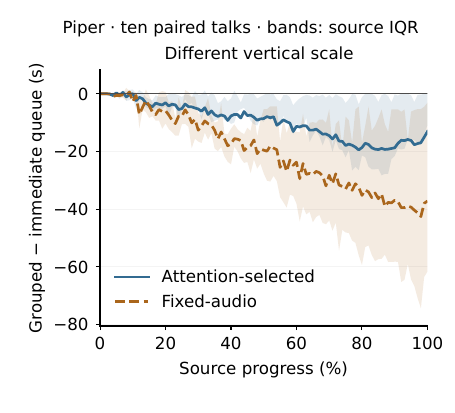}
\caption{Change in ready-audio queue during source arrival. Curves show mean paired changes under four-second rather than immediate grouping; bands show the interquartile range across ten talks. Negative values mean less already-generated audio waiting to play, not earlier word delivery. The horizontal axis is source progress; the panels use visibly different vertical scales. Post-source tails remain in the completion tables.}
\label{fig:queue}
\end{figure*}

\Needspace{12\baselineskip}
\section{Ready-Audio Queue during Source Arrival}
\label{app:queue}
Final completion does not describe the queue during source arrival. For positive-duration part $j$, let $r_j$, $p_j$, and $C_j$ be readiness, playback start, and playback end. At source-start time $t$, ready but unplayed audio duration is
\begin{equation}
 Q(t)=\sum_{j:r_j\leq t}\max\{0,C_j-\max(p_j,t)\}.
 \label{eq:queue}
\end{equation}
This includes the unplayed remainder of a currently playing part, but excludes audio not yet generated. A low queue is therefore not evidence of prompt or faithful translation.

For source duration $S_s$ and progress $u\in[0,1]$, we compare the same source-clock time:
\begin{equation}
 \Delta Q_s(u)=Q_{s,4}(uS_s)-Q_{s,0}(uS_s).
\end{equation}
Figure~\ref{fig:queue} shows equal-talk means on a 101-point source-progress grid. Bands are the 25th--75th source percentiles with linear quantile interpolation, not confidence intervals. Queue values are reconstructed from readiness jumps and playback start/end points, not interpolated between sparse source observations. No smoothing is used. The figure stops at source end; full post-source draining remains in the completion results and archived queue paths.

\end{document}